\documentclass{article}

\usepackage[preprint]{neurips_2025}

\usepackage[utf8]{inputenc} 
\usepackage[T1]{fontenc}    
\usepackage{hyperref}       
\usepackage{url}            
\usepackage{booktabs}       
\usepackage{amsfonts}       
\usepackage{nicefrac}       
\usepackage{microtype}      
\usepackage{xcolor}         
\usepackage{epsfig}
\usepackage{etoolbox}
\usepackage{float}
\usepackage{graphicx}
\usepackage{amsmath}
\usepackage{color}
\usepackage{amssymb}
\usepackage{wrapfig,setspace}
\usepackage{tabu}                     
\usepackage{multirow}                 
\usepackage{multicol}                 
\usepackage{multirow}                
\usepackage{float}                    
\usepackage{makecell}                 
\usepackage{booktabs}                 
\usepackage{pifont}
\usepackage{diagbox}
\usepackage{csquotes} 
\usepackage{multicol}
\usepackage{minitoc}
\usepackage{tocloft}
\usepackage{url}
\usepackage{hyperref}       
\usepackage{url}            
\usepackage{booktabs}       
\usepackage{amsfonts}       
\usepackage{nicefrac}       
\usepackage{microtype}      
\usepackage{xcolor}         
\usepackage{graphicx}
\usepackage{amsmath}
\usepackage{multirow}
\usepackage{amsmath}
\usepackage{colortbl}
\usepackage{pifont}

\title{\raisebox{-0.16cm}{\includegraphics[scale=0.06]{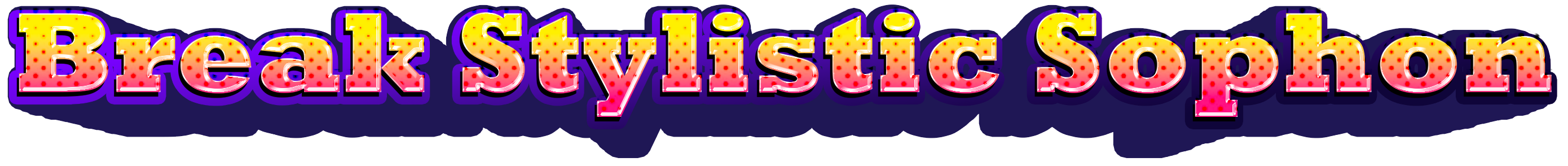}}: Are We Really Meant to Confine the Imagination in Style Transfer?}

\author{%
  Gary Song Yan \\
  ISAI Lab\\
  Xi'an Institute of High-tech\\
  Xi'an, China \\
  \texttt{gary\_144@outlook.com} \\
  \And
  Yusen Zhang \\
  ISAI Lab\\
  Xi'an Institute of High-tech\\
  Xi'an, China \\
  \texttt{ysen1101@163.com} \\
  \And
  Jinyu Zhao \\
  ISAI Lab\\
  Xi'an Institute of High-tech\\
  Xi'an, China \\
  \texttt{369360770@qq.com} \\
  \And
  Hao Zhang \\
  Department of Basic Courses\\
  Xi'an Institute of High-tech\\
  Xi'an, China \\
  \texttt{zh01020938@163.com} \\
  \And
  Zhangping Yang \\
  ISAI Lab\\
  Xi'an Institute of High-tech\\
  Xi'an, China \\
  \texttt{akame1027@163.com} \\
  \And
  Guanye Xiong \\
  ISAI Lab\\
  Xi'an Institute of High-tech\\
  Xi'an, China \\
  \texttt{13350433677@163.com} \\
  \And
  Yanfei Liu \\
  Department of Basic Courses\\
  Xi'an Institute of High-tech\\
  Xi'an, China \\
  \texttt{bbmcu@126.com} \\
  \And
  Tao Zhang \\
  Department of Computer Science\\
 Huazhong University of Science and Technology\\
  Wuhan, China \\
  \texttt{zhangtao\_2023@hust.edu.cn} \\
  \And
  Yujie He \\
 ISAI Lab\\
 Xi'an Institute of High-tech\\
 Xi'an, China \\
 \texttt{ksy5201314@163.com} \\
 \And
 Siyuan Tian \\
 College of Science\\
 National University of Defence Technology\\
 Changsha, China \\
 \texttt{tiansy@nudt.edu.cn} \\
 \And
 Yao Gou \\
 Intelligent Game and Decision Lab\\
 Beijing, China \\
 \texttt{gouayao@163.com} \\
 \And
 Min Li\thanks{Corresponding author}  \\
ISAI Lab\\
Xi'an Institute of High-tech\\
Xi'an, China \\
\texttt{proflimin@163.com} \\
}

\begin{document}

	\maketitle
	\begin{figure}[h]
		\centering
		\includegraphics[width=\linewidth]{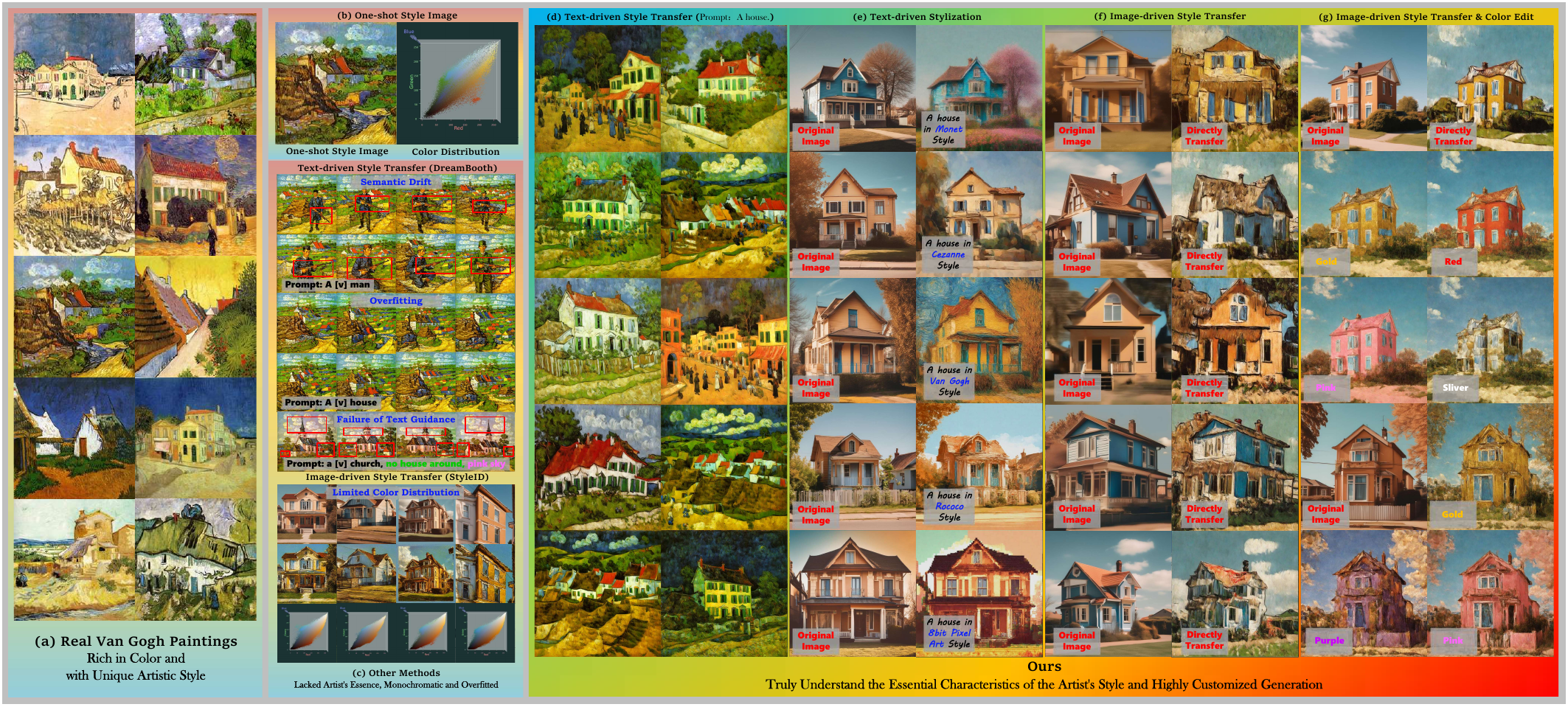}
		\caption{We found that existing image style transfer methods based on a single style image (b) either suffer from overfitting and semantic drift when performing text-driven style transfer or merely achieve texture blending rather than truly learning the artist's style at the level of artistic creation during image-driven style transfer (c). Moreover, there is currently a lack of a unified framework for addressing the various issues in style transfer. For genuine artworks (a), the colors are never confined to just one piece (b). Therefore, this paper designs a unified model that can learn the artist's creative style in the true sense and achieves style transfer results that are indistinguishable from the artist's creative style (d), (e) and (f). Meanwhile, the model also realizes color editing during the style transfer process for the first time (g).}
		\label{fig:intro}
	\end{figure}
	
	
	\begin{abstract}
		In the realm of image style transfer, existing algorithms relying on single reference style images encounter formidable challenges, such as severe semantic drift, overfitting, color limitations, and a lack of a unified framework. These issues impede the generation of high quality, diverse, and semantically accurate images. In this pioneering study, we introduce \textit{StyleWallfacer}, a groundbreaking unified training and inference framework, which not only addresses various issues encountered in the style transfer process of traditional methods but also unifies the framework for different tasks. This framework is designed to revolutionize the field by enabling artist level style transfer and text driven stylization. First, we propose a semantic-based style injection method that uses BLIP to generate text descriptions strictly aligned with the semantics of the style image in CLIP space. By leveraging a large language model to remove style-related descriptions from these descriptions, we create a semantic gap. This gap is then used to fine-tune the model, enabling efficient and drift-free injection of style knowledge. Second, we propose a data augmentation strategy based on human feedback, incorporating high-quality samples generated early in the fine-tuning process into the training set to facilitate progressive learning and significantly reduce its overfitting. Finally, we design a training-free triple diffusion process using the fine-tuned model, which manipulates the features of self-attention layers in a manner similar to the cross-attention mechanism. Specifically, in the generation process, the key and value of the content-related process are replaced with those of the style-related process to inject style while maintaining text control over the model. We also introduce query preservation to mitigate disruptions to the original content. Under such a design, we have achieved high-quality image-driven style transfer and text-driven stylization, delivering artist-level style transfer results while preserving the original image content. Moreover, we achieve image color editing during the style transfer process for the first time, further pushing the boundaries of controllable image generation and editing technologies and breaking the limitations imposed by reference images on style transfer. Our experimental results demonstrate that our proposed method outperforms state-of-the-art methods. 
	\end{abstract}
	
	\section{Introduction}
	Art encapsulates human civilization's essence, epitomizing our imagination and creativity, and has yielded innumerable masterpieces. Online, you may encounter a painting that profoundly affects you, yet you may find it hard to describe the artist's unique style or locate more similar works. This highlights a key issue in image generation: style transfer.
	
	
	Recently, numerous excellent works have conducted research on this issue, which are mainly divided into three categories: text-driven style transfer \cite{rout2025rbmodulation,DBLP:conf/nips/SohnJBLRKCLERHE23}, image-driven style transfer \cite{DBLP:conf/cvpr/ChungHH24,DBLP:conf/iccv/LiuLHLWLSLD21,DBLP:conf/iccv/HuangB17} and text-driven stylization \cite{jiang2024artist,DBLP:conf/cvpr/BrooksHE23,Tumanyan_2023_CVPR}. The mainstream approach of image-driven style transfer is to decouple the style and content information of a reference style image, and then inject the style information as an additional condition into the model's generation process \cite{DBLP:conf/cvpr/ChenTH24,DBLP:journals/corr/abs-2407-00788}. This enables the model to generate new content that is similar to the reference style image in terms of texture and color. Alternatively, a unique identifier can be used to characterize the style of the style image, and the model can be fine-tuned to learn new stylistic knowledge for text-driven style transfer \cite{DBLP:conf/cvpr/RuizLJPRA23,DBLP:conf/iccv/HanLZMMY23}. This allows the model to recognize and generate corresponding style images using the identifier. For text-driven stylization, most methods involve blending the pre-trained model's prior style knowledge with the texture of the target image to achieve the final style transfer result \cite{jiang2024artist}. However, as shown in Figure~\ref{fig:intro}~(c), these models generally suffer from the following issues:
	
	\textbf{Limited color domain}: Both the style-content disentanglement-based and identifier-based fine-tuning approaches commonly face the problem of a restricted color domain in the generated images. Specifically, the color distribution of the generated images is entirely consistent with that of the single reference style image. For example, in the case of Van Gogh's paintings \cite{DBLP:conf/cvpr/OjhaLLELS021} as shown in Figure~\ref{fig:intro}~(a), great artists are by no means limited to the color palette of a single artwork in Figure~\ref{fig:intro}~(b). Therefore, such generation results are unreasonable. For more detailed visualization results and discussions, please refer to Appendix~\ref{LCD}.
	
	\textbf{Failure of text guidance}: Due to the architectural flaws in the style-content disentanglement-based methods and the mismatch between text and image in the style information injection of the identifier-based fine-tuning methods, models exhibit significant semantic drift, which refers to the phenomenon of inconsistency or deviation in semantics between the generated image and the input text prompt in the T2I model. This not only leads to chaotic generation but also results in the loss of the model's ability to handle complex text prompts. For more detailed visualization results and discussions, please refer to Appendix~\ref{FTG}.
	
	\textbf{Risk of overfitting }: Due to the extremely limited number of training samples, traditional approaches are generally prone to overfitting. This results in a loss of structural diversity in the generated content. For more detailed visualization results and discussions, please refer to Appendix~\ref{RO}.
	
	\textbf{Lack of a unified framework}: Due to the significant differences between various style transfer tasks, most existing style transfer methods are only capable of handling one specific task, and there is a lack of a unified framework to integrate these tasks.
	
\begin{figure}[htbp]
		\centering
		\includegraphics[width=\linewidth]{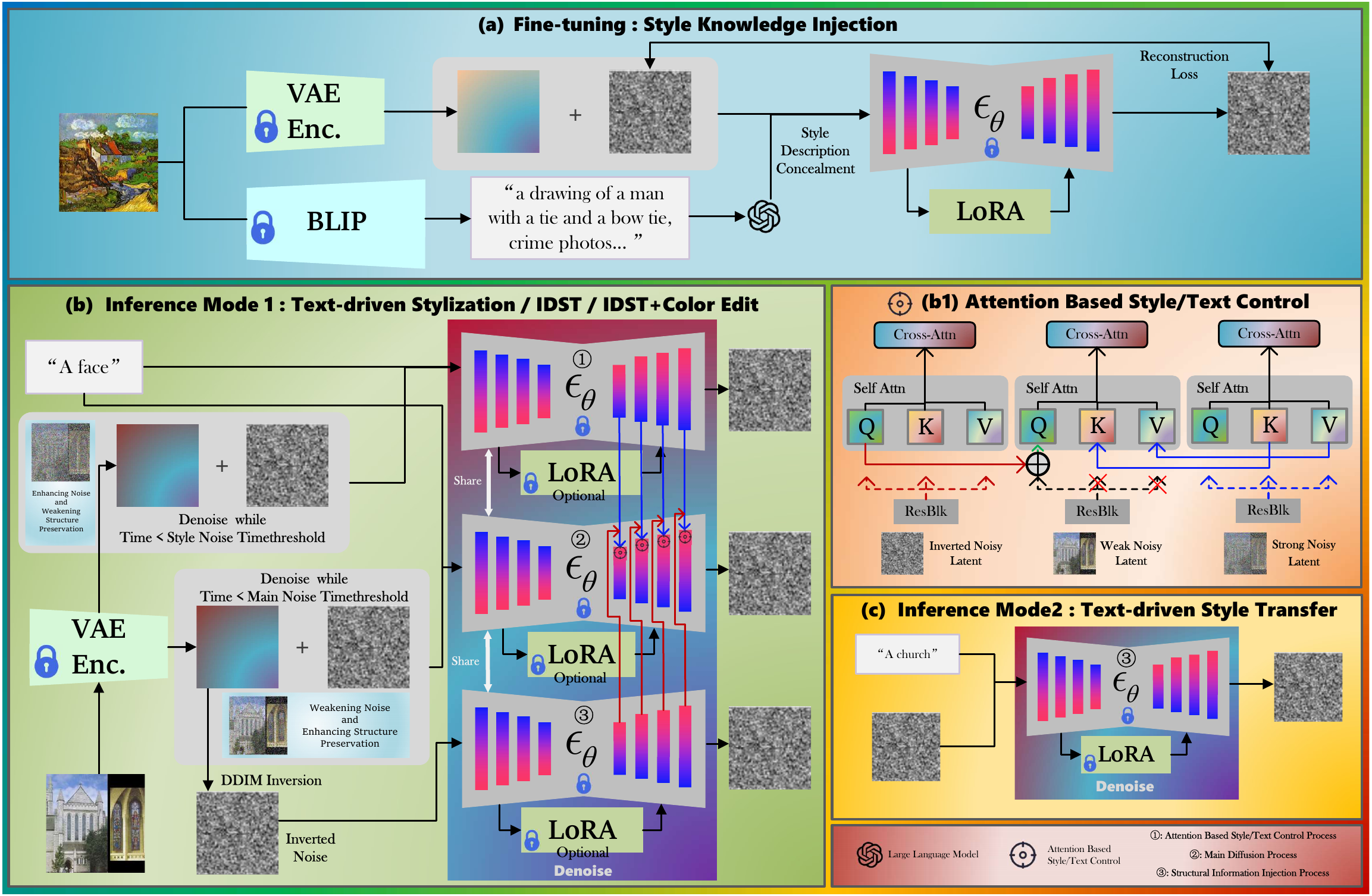}
		\vspace{-11pt}
		\caption{\textbf{Illustration of the \textit{StyleWallfacer} Framework.} In the fine-tuning stage (a), we use a semantic-based style knowledge injection method with human feedback (see Figure~\ref{fig:ft}) fine-tuning to help the model learn the style knowledge of a single image, obtaining fine-tuned style LoRA weights. This enables powerful text-driven style transfer (c). In the inference stage (b), we design a triple training-free diffusion pipeline (denoted as \ding{172}, \ding{173}, \ding{174}). It use the diffusion denoising process with a smaller threshold \( t_s^s \) as the main process and extract \textit{Key} \(\mathbf{K}_t^l\) and \textit{Value} \(\mathbf{V}_t^l\) from the process with a larger threshold \( t_s^l \) to guide the main process in style and text. Additionally, we use the DDIM inversion latent's denoising diffusion process as the third guiding process, extracting its \textit{Query} \(\mathbf{Q}_t^i\) to inject into the main process, achieving artist-level image-driven style transfer, text-driven stylization and color edit (b1). For more detailed introductions to the pipelines, please refer to Appendix~\ref{inf}.} 
		\label{fig:stru}
	\end{figure}

	These problems, much like the "sophons" in "The Three-Body Problem" \cite{TheThreeBodyProblem} that restrict human technological progress, limit people's imagination for style transfer. In fact, truly good style transfer enables the model to learn and imitate the artist's creative style, rather than mechanically copying the textures of reference images, thus achieving true artistic creation. Just as in "The Three-Body Problem", humanity uses the "Wallfacer Plan" to break the technological blockade imposed by the "sophons", to break the limitations of "sophons" in style transfer, this paper proposes a novel unified style transfer framework, called \textit{StyleWallfacer}, which consists of three main components: 
	
	Firstly, a style knowledge injection method based on semantic differences is proposed (Figure~\ref{fig:stru} (a)). By using BLIP \cite{DBLP:conf/icml/0001LXH22} to generate text descriptions that are strictly aligned with the target style image in CLIP space \cite{DBLP:conf/icml/RadfordKHRGASAM21}, and then leveraging LLM to remove the style-related descriptions, a semantic gap is created. This gap allows the model to maintain its prior knowledge as much as possible during training, focusing solely on learning the style information. As a result, the model captures the most fundamental stylistic elements of the style image (e.g., the artist's brushstrokes). As shown in Figure~\ref{fig:intro}~(d), this not only enables the generation of new samples with rich and diverse colors but also preserves the model's ability to handle complex text prompts. 
	
	Secondly, a progressive learning method based on human feedback (HF) is employed (Figure~\ref{fig:ft}). At the beginning of model training, the model is trained using a single sample. During the training process, users are allowed to select high-quality samples generated by the model and add them to the training set. This effectively expands the single-sample dataset and significantly mitigates overfitting of the model. 
	
	Thirdly, we propose a brand-new training-free triple diffusion “style-structure” diffusion process (Figure~\ref{fig:stru} (b) and (b1)). It explores the impact of different noise thresholds on the model's generation effects by using the diffusion process with a smaller noise threshold as the main process to preserve the content information of the original image, and employing the diffusion process with a larger noise threshold as the style guidance process. Meanwhile, the \textit{Key} and \textit{Value} from the self-attention layer during this process are extracted to replace the \textit{Key} and \textit{Value} in the main diffusion process and obtain the initial noise of the style image to be transferred through DDIM inversion \cite{DBLP:conf/iclr/SongME21}. The \textit{Query} from the diffusion process of the inverted noise is extracted and fused with the \textit{Query} in the main diffusion process, serving as a structural guidance for the main diffusion process. Meanwhile, the pre-trained style LoRA \cite{DBLP:conf/iclr/HuSWALWWC22} is used as a style guide to direct the model to conduct image-driven style transfer. This approach thus achieves the artist-level style transfer results as shown in Figure~\ref{fig:intro}~(e) and (f). During the generation process, text prompt is employed as a condition, and in combination with the aforementioned structure, it also enables color editing of the model during the style transfer process as shown in Figure~\ref{fig:intro}~(g).
	
	Our main contributions are summarized as follows:
	
	\textbf{(1)} We propose the first unified style transfer framework that simultaneously achieves high-quality style transfer. Meanwhile, for the first time, it enables text-based color editing during the style transfer process.
	
	\textbf{(2)} We propose a style knowledge injection method based on semantic differences, which achieves efficient style knowledge injection without affecting the model's semantic space and suppresses semantic confusion during the style injection process.
	
	\textbf{(3)} We propose a progressive learning method based on human feedback for few-shot datasets, which alleviates the model overfitting caused by insufficient data and significantly improves the generation quality after model training.
	
	\textbf{(4)} We propose a novel training-free triple diffusion process that achieves artist-level style transfer results while retaining the control ability of text prompts over the generation results, and for the first time enables color editing during the style transfer process.
	
	\textbf{(5)} Our experiments demonstrate that the proposed method in this paper addresses many issues encountered by traditional methods during style transfer, achieving artist-level style generation results rather than merely texture blending, and delivering state-of-the-art performance.
	
	\section{\textit{StyleWallfacer}}
	\label{headings}
	\subsection{Overall Architecture of \textit{StyleWallfacer}}
	As shown in Figure~\ref{fig:stru}, \textit{StyleWallfacer} mainly consists of two parts: First is the semantic-based style learning strategy, which aims to guide the model to learn the most essential style features in artworks based on the semantic differences between images and their text descriptions during the model fine-tuning process, truly helping the model understand the artist's creative style. It also employs a data augmentation method based on human feedback to suppress overfitting when the model is fine-tuned on a single image, thereby achieving realistic text-driven style transfer. 
	
	The second part is the training-free triple diffusion process, which is designed using the previously fine-tuned LoRA weights. This section comprises three newly designed pipelines tailored to address different style transfer problems. By adjusting the self-attention layers of three denoising networks that share weights (denoted as \includegraphics[width=0.02\linewidth]{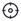}), it achieves high-quality style control. This results in artist-level style transfer and, for the first time, enables text prompts to control image colors during the style transfer process,  solving the traditional method's shortcomings of monochromatic colors, simple textures, and lack of text control when transferring styles based on a single image.
	\subsection{Semantic-based Style Learning Strategy}
	\label{SBSLS}

	The semantic-based style learning strategy primarily aims to fine-tune text-to-image (T2I) models using their native “language” to enhance their comprehension of the knowledge humans intend them to learn during the fine-tuning process. Taking Stable Diffusion as an example, there is a significant discrepancy between the image semantics understood by the pre-trained CLIP and the intuitive human understanding of image semantics. Therefore, to better “communicate” with the pre-trained T2I model during fine-tuning, this paper employs a method of reverse-engineering the semantic information of image \( \mathbf{I} \) in the CLIP space through BLIP \cite{DBLP:conf/icml/0001LXH22}:
	\begin{equation}
		\begin{aligned}
			\mathbf{T}_{\text{CLIP}} =\text{BLIP}\left ( \mathbf{I} \right ) 
		\end{aligned}
	\end{equation}
	where  \(\mathbf{T}_{\text{CLIP}}\) denotes the image prompt derived through BLIP.
	
	Although such methods enable us to obtain the semantic information corresponding to an image in the CLIP space, this text description cannot be directly employed in the fine-tuning process. This is because the description  \(\mathbf{T}_{\text{CLIP}}\) encompasses all information pertaining to the image, including content, style, and other details understood by CLIP. Utilizing this comprehensive description for fine-tuning still results in the model's inability to comprehend human fine-tuning intentions, thereby preventing it from learning the stylistic information of the dataset.
	
	Therefore, our \textit{StyleWallfacer} transforms \(\mathbf{T}_{\text{CLIP}}\) by creating an semantic discrepancy among descriptions. By incorporating a large language model to perform subtle semantic edits on \(\mathbf{T}_{\text{CLIP}}\), descriptions related to image style are selectively removed: 
	\begin{equation}
		\begin{aligned}
			\mathbf{T}_{w/oS} =\text{LLM}\left ( \mathbf{T}_{\text{CLIP}}\right ) 
		\end{aligned}
	\end{equation}
	where \(\mathbf{T}_{w/oS}\) denotes the text description after removing the style information, and LLM stands for large language model.
	
	After such processing, we obtain the image \(\mathbf{I}\) and its corresponding text description \(\mathbf{T}_{w/oS}\) in the CLIP space, from which stylistic descriptions have been removed. As shown in Figure~\ref{fig:stru}~(a), fine-tuning a pre-trained T2I model using these image-text pairs enables it to focus more effectively on understanding stylistic information, thereby circumventing unnecessary semantic drift.
	
	\subsection{Training-free Triple Diffusion Process}
	After fine-tuning, the model has essentially learned the most fundamental style knowledge from the reference style image. Therefore, how to activate this knowledge so that it can be utilized for image-driven style transfer has become an extremely critical issue.
	
	Unlike traditional one-shot style transfer algorithms that require the reference style image as input during style transfer, we aim to rely solely on the pre-trained style LoRA obtained in Section~\ref{SBSLS} for style transfer. Therefore, we cannot adopt a method similar to StyleID \cite{DBLP:conf/cvpr/ChungHH24} to manipulate the features in the self-attention layer as if they were cross-attention features, with the features from the style image \( \mathbf{I}_s \) serving as the condition for style injection.
	
	\begin{figure}[h]
		\centering
		\includegraphics[width=0.8\linewidth]{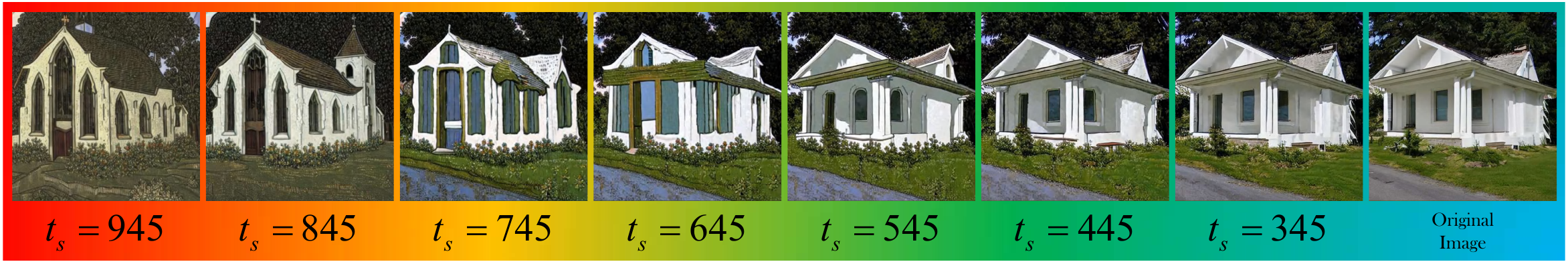}
		\caption{\textbf{Illustration of the Impact of Noise Schedule Threshold \( t_s \) on Model Generation Results.}} 
		\label{fig:t}
	\end{figure}
	
	However, as shown in Figure~\ref{fig:t}, we observe that when initializing the noisy latent \( \mathbf{X}_0 \) with the original image and using the U-Net to denoise it, the larger the noise schedule threshold \( t_s \), the more stylized the generated image will be, losing the original image's content information and retaining only its basic semantics. Conversely, the smaller the noise schedule threshold \( t_s \), the more the model's generation tends to preserve the original image's content information, while reducing the diversity and stylization in the generation process.
	
	Therefore, we contemplate: Is it possible to fully leverage this characteristic by employing a diffusion process with a smaller \( t_s \) as the main diffusion process, and using a diffusion process with a larger \( t_s \) as the stylistic guiding process? Meanwhile, we can utilize the inverted latent obtained through DDIM inversion as the noisy latent for the third diffusion process, and harness the residual information from its denoising process as content guidance. In this way, we aim to achieve high-quality style transfer results while preserving the image content.
	
	To this end, as shown in Figure~\ref{fig:stru}~(b), we first use the VAE encoder to transform the image \( \mathbf{I}_c \) to be transferred from the pixel space to the latent space, obtaining \( \mathbf{F}_0 \). By setting a larger noise schedule threshold \( t_s^{l} \), we add noise to \( \mathbf{F}_0 \) (at \( t=0 \)) to obtain \( \mathbf{F}_l \) (at \( t=t_s^{l} \)). Similarly, by using a smaller noise schedule threshold \( t_s^{s} \) , we obtain \( \mathbf{F}_s \) (at \( t=t_s^{s} \)). Additionally, we use DDIM inversion to invert \( \mathbf{F}_0 \) to Gaussian noise \( \mathbf{F}_{i} \) (at \( t=T \)). Then, using the same denoising U-Net, we denoise \( \mathbf{F}_s \), \( \mathbf{F}_l \), and \( \mathbf{F}_i \) respectively.  As shown in Figure~\ref{fig:stru}~(b1), during the entire denoise process of latent \( \mathbf{F}_s \), we transfer \( \mathbf{F}_s \) to \( \mathbf{F}_l \) by injecting the \textit{Key} \( \mathbf{K}_t^l \) and \textit{Value} \( \mathbf{V}_t^l \) collected from \( \mathbf{F}_l \) into the self-attention layer, instead of the original \textit{Key} \( \mathbf{K}_t^s \) and \textit{Value} \( \mathbf{V}_t^s \). However, merely implementing this substitution can result in content disruption, as the content of the \( \mathbf{F}_s \) representation would be progressively altered with the changes in the attended values. 
	
	Consequently, we propose a query preservation mechanism to retain the original content. Simply, as shown in Figure~\ref{fig:stru}~(b1), we fuse the \textit{Query} \( \mathbf{Q}_t^i \) of DDIM inverted latent \( \mathbf{F}_i \) with the original \textit{Query} \( \mathbf{Q}_t^s\) to get \textit{Query} \( \mathbf{Q}_t^f\) and inject it to the main denoise process instead of the original \textit{Query} \( \mathbf{Q}_t^s\). These style injection, query preservation and structural residual injection processes at time step t are expressed as follows:
	\begin{flalign}
		&\mathbf{Q}_{t}^{f}=\beta\mathbf{Q}_t^i+(1-\beta)\mathbf{Q}_t^s ,\\
		&\phi_{\mathrm{out}}^{l}=\mathrm{Attn}(\mathbf{Q}_{t}^{f},\mathbf{K}_{t}^{l},\mathbf{V}_{t}^{l}),
	\end{flalign}
	where $\beta\in[0,1]$. $\mu(\cdot )$, $\sigma(\cdot )$ and $\phi_{\mathrm{out}}^{l}$ denote channel-wise mean, standard deviation and the result of self-attention calculation after replacement, respectively. In addition, we apply these operations on the decoder of U-net in SD. We also highlight that the proposed method can adjust the degree of style transfer by changing noise schedule threshold \( t_s^{l} \) and \( t_s^{s} \). Specifically, lower \( t_s^{l} \) and \( t_s^{s} \) maintains more content, while higher \( t_s^{l} \) and \( t_s^{s} \) strengthens effects of style transfer.
	
	\subsection{Data Augmentation for Small Scale Datasets Based on Human Feedback}
    Although this paper proposes a more robust style knowledge injection method than DreamBooth \cite{DBLP:conf/cvpr/RuizLJPRA23} in Section~\ref{SBSLS}, fine-tuning models with a single sample remains challenging. Therefore, inspired by human feedback reinforcement learning (HFRL) \cite{shen2025exploringdatascalingtrends}, this paper proposes a human feedback-based data augmentation method for small-scale datasets to compensate for dataset insufficiency and mitigate overfitting.
	
	\begin{figure}[h]
		\centering
		\includegraphics[width=0.85\linewidth]{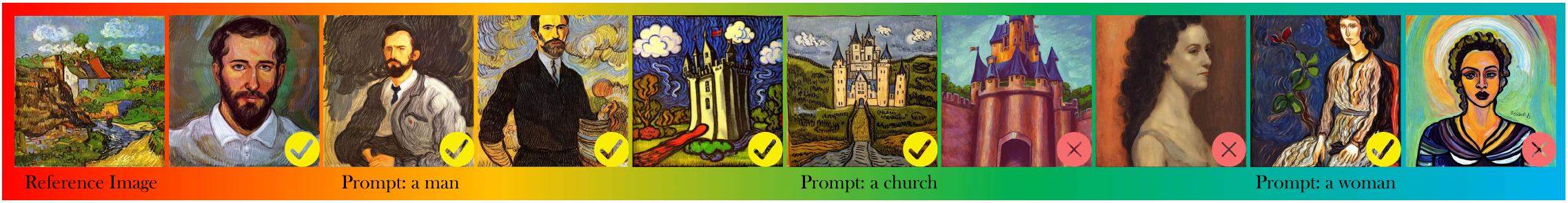}
		\caption{\textbf{Illustration of the Best Generation Results When Fine-tuning the Model Directly with a Single Image.}}
		\label{fig:gab}
	\end{figure}
	Specifically, when the model is first trained on a single style image, as shown in Figure~\ref{fig:gab}, although the injection of style knowledge does not generalize well to all the prior knowledge of the model in the early stages of training, and some of the generated results do not match the reference style consistently, resulting in an asynchronous phenomenon in the injection of style knowledge. However, there are still many excellent samples in the model's generated results. The reason for the emergence of these samples is that the prior knowledge represented by these samples is similar to the style image used in training in the CLIP space. This makes it easier for the model to transfer style knowledge to these pieces of knowledge during training. Therefore, it is possible to select samples that meet the style requirements from the large number of generated samples and add them to the training set for further fine-tuning of the model. 
	
	\begin{figure}[h]
		\centering
		\includegraphics[width=0.8\linewidth]{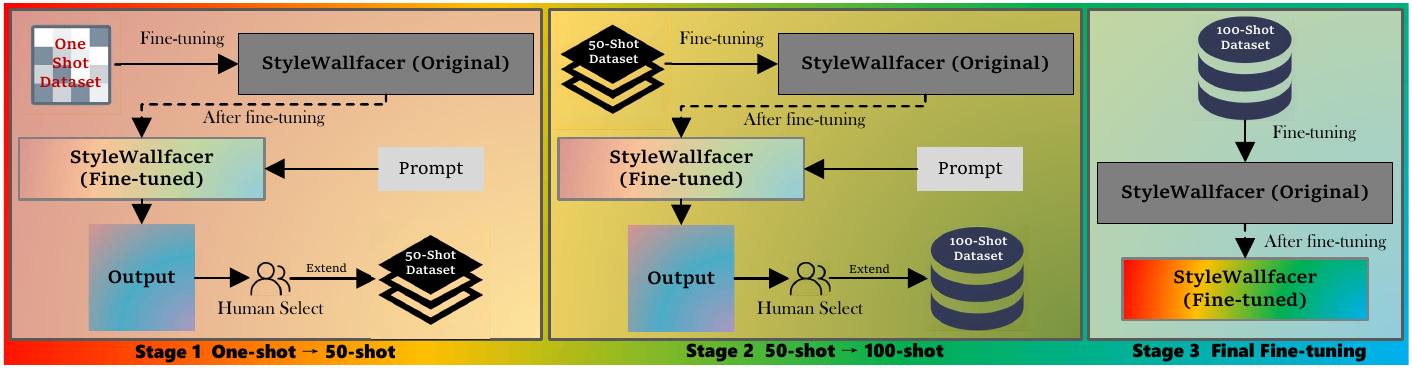}
		\caption{\textbf{Illustration of the Small Scale Datasets Augmentation Method Based on Human Feedback.}}
		\label{fig:ft}
	\end{figure}
	
	To this end, as shown in Figure~\ref{fig:ft}, we divide the model's fine-tuning process into three stages. In the first stage, which is the single-sample fine-tuning stage, we generate a large number of new samples using a text prompt that reflects the basic semantics of the reference image after the model has been trained. We then manually select the 50 samples that best match the stylistic features of the reference image and add them to the training set for the second stage of fine-tuning. In the second stage of fine-tuning, the basic idea is similar to the first stage. We expand the training set from 50 to 100 images. Finally, we fine-tune the model using these 100 samples to obtain the final style LoRA.
	
	Through this data augmentation strategy, the overfitting phenomenon of the model during the fine-tuning process is greatly alleviated. The fine-tuned model is able to generate more diverse results and generalize the style knowledge to all the prior knowledge of the model, rather than being limited to a single sample.
	
	\vspace{-5pt}
	\section{Experiments}
	\vspace{-5pt}
	\label{others}
	\subsection{Experimental Setting}
	\textbf{Baselines}~~~Our baseline list in one-shot text-driven style transfer includes DreamBooth \cite{DBLP:conf/cvpr/RuizLJPRA23}, a LoRA \cite{DBLP:conf/iclr/HuSWALWWC22} version of DreamBooth, Textual Inversion \cite{DBLP:conf/iclr/GalAAPBCC23}, and SVDiff \cite{DBLP:conf/iccv/HanLZMMY23}. For the text-driven stylization task, the baselines we selected include Artist \cite{jiang2024artist}, InstructPix2Pix \cite{DBLP:conf/cvpr/BrooksHE23} , and Plug-and-play (PnP) \cite{Tumanyan_2023_CVPR}. And baseline list in one-shot image-driven style transfer includes StyleID \cite{DBLP:conf/cvpr/ChungHH24}, AdaAttn \cite{DBLP:conf/iccv/LiuLHLWLSLD21}, AdaIN \cite{DBLP:conf/iccv/HuangB17}, AesPA-Net \cite{DBLP:conf/iccv/HongJLAKLKUB23} and InstantStyle-Plus \cite{DBLP:journals/corr/abs-2407-00788}.
	
	\textbf{Datasets}~~~We selected one image from each of the three widely used 10-shot datasets, including landscapes \cite{XiaogangWang_XiaoouTang_2009}, Van Gogh houses \cite{DBLP:conf/cvpr/OjhaLLELS021}, and watercolor dogs \cite{DBLP:journals/corr/abs-2306-00763}, to form our one-shot datasets, in order to quantitatively evaluate the proposed method from a better perspective. To test our model, we first used FLUX \cite{flux2024} to generate 1,000 images of houses, 1,000 images of dogs and 1,000 images of mountains based on the prompts "a photo of a house", "a photo of a dog" and "a photo of a moutain", respectively. These images served as the style-free images to be transferred.
	
	\textbf{Metric}~~~For image style similarity, we compute CLIP-FID \cite{DBLP:conf/cvpr/Parmar0Z22}, CLIP-I score, CLIP-T score and DINO score \cite{DBLP:conf/iclr/0097LL000NS23} between 1,000 samples with the full few-shot datasets. For image content similarity, we compute the LPIPS \cite{DBLP:conf/cvpr/Parmar0Z22} between 1,000 samples and the source image to evaluate the content similarity between the style-transferred images and the original images. Intra-clustered LPIPS \cite{Ojha_Li_Lu_Efros_JaeLee_Shechtman_Zhang_2021,DBLP:conf/cvpr/ZhangIESW18} of 1,000 samples is also reported as a standalone diversity metric.
	
	\textbf{Detail}~~~For other details of the experiments and \textit{StyleWallfacer}, please refer to Appendix~\ref{EXP}.\vspace{-5pt}
	
	\subsection{Qualitative Comparison}
	\begin{figure}[t]
		\label{re}
		\centering
		\includegraphics[width=\textwidth]{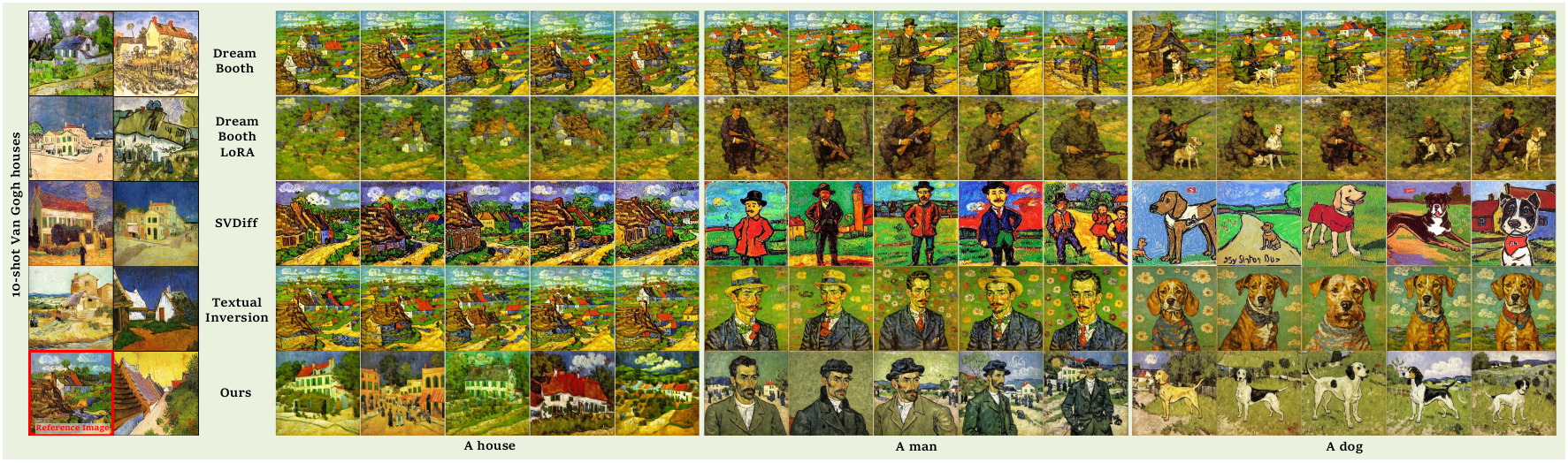}
		\vspace{-14pt}
		\caption{\textbf{Qualitative Comparison of Text-driven Style Transfer Results on Van Gogh houses Dataset Using Different Methods.} Due to page limitations, we have placed some of the experimental results in Appendix~\ref{mr1}.}
		\label{re}
	\end{figure}
	
	\begin{figure}[t]
		\label{vg3}
		\centering
		\includegraphics[width=\textwidth]{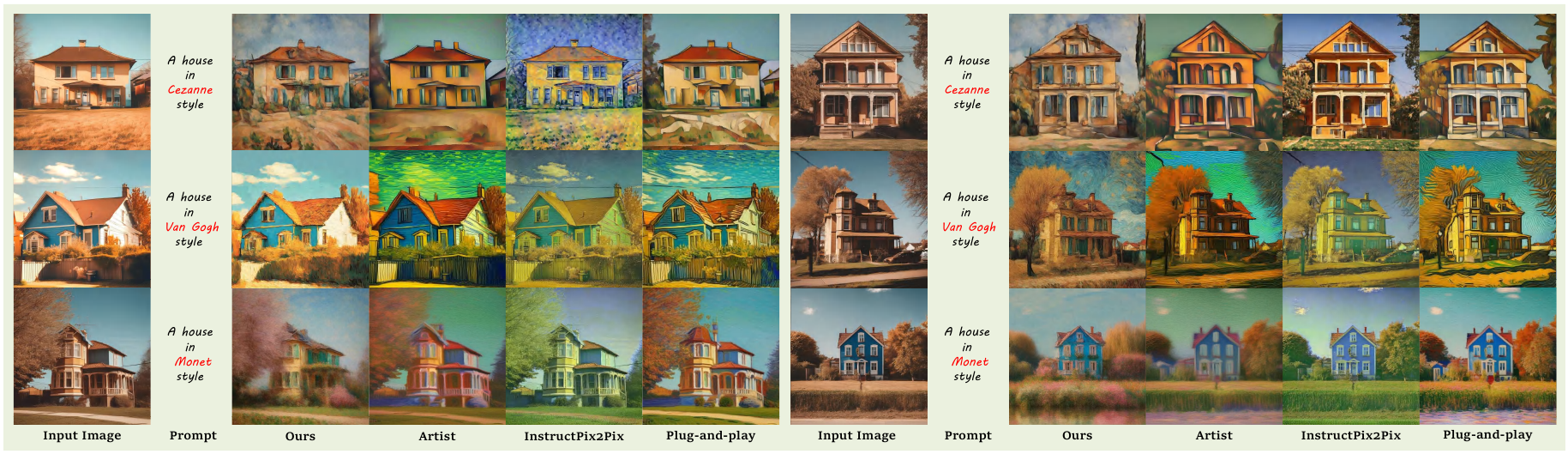}
		\vspace{-14pt}
		\caption{\textbf{Qualitative Comparison of Text-driven Stylization Results Using Different Methods.} Due to page limitations, we have placed some of the experimental results in Appendix~\ref{mr4}.}
		\label{vg3}
	\end{figure}
	\textbf{One-shot Text-driven Style Transfer Experimental Qualitative Results.}
	As depicted in Figure~\ref{re}, \textit{StyleWallfacer} outperforms other methods in generating diverse and semantically accurate images. Unlike other methods that suffer from overfitting and semantic drift when trained on single-style images, \textit{StyleWallfacer} employs multi-stage progressive learning with human feedback to reduce overfitting and enhance diversity. It also avoids identifiers for style injection, minimizing semantic drift and enabling precise style generation based on prompts.
	
	\textbf{Text-driven Stylization Experimental Qualitative Results.}
	As shown in Figure~\ref{vg3}, other methods, except \textit{StyleWallfacer}, although have completed the task of style transfer, the results obtained after the transfer are far from the authentic style of the painter and fall short of the expected level. However, \textit{StyleWallfacer} has achieved the best balance between image style transfer and content preservation. The images after style transfer not only closely match the painter's authentic style but also feature finer details and a high degree of fidelity to the original image content.
	
	\textbf{One-shot Image-driven Style Transfer Experimental Qualitative Results.}
	\begin{figure*}[t]
		\label{vg2}
		\centering
		\includegraphics[width=\textwidth]{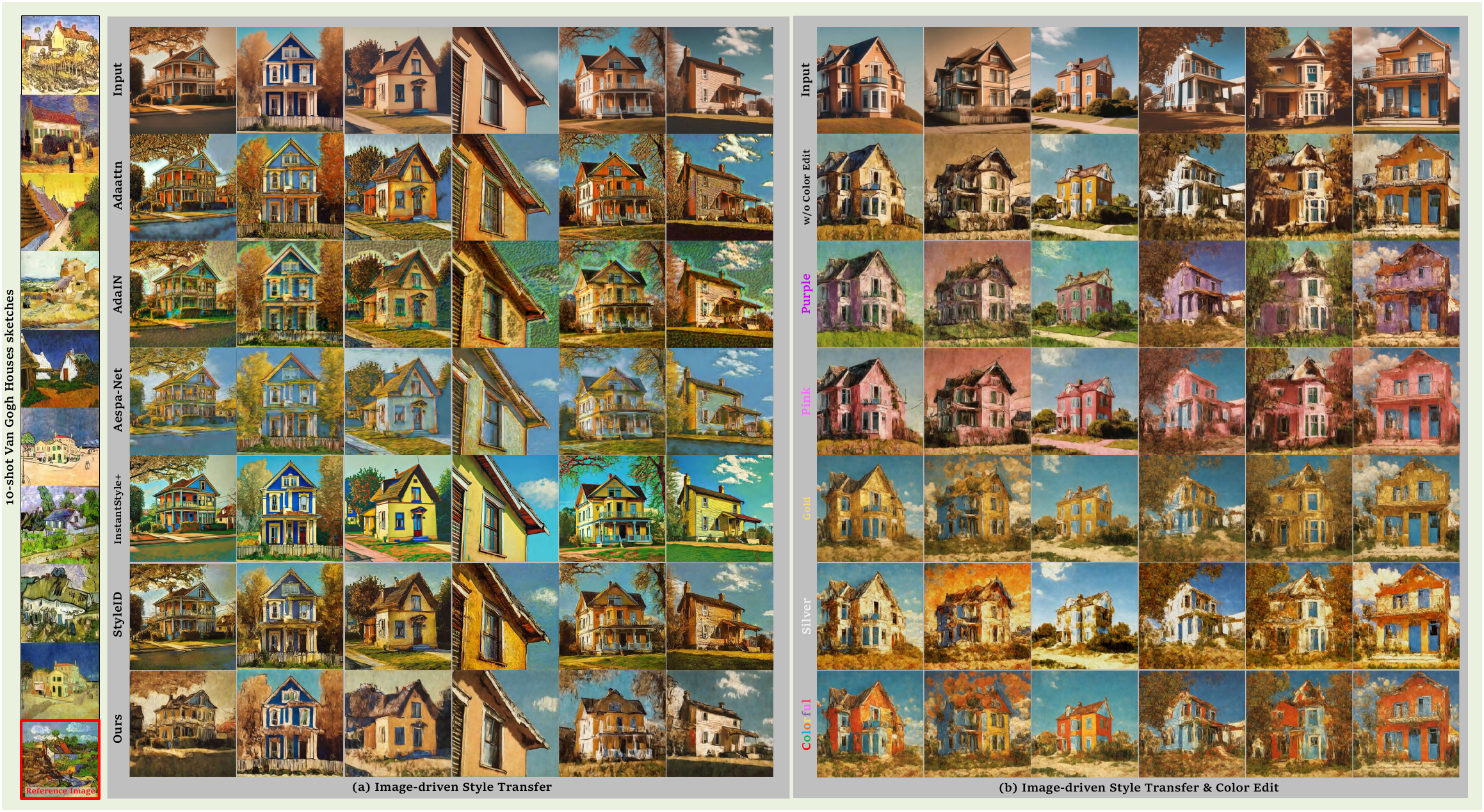}
		\vspace{-14pt}
		\caption{\textbf{Qualitative Comparison of Image-driven Style Transfer and Color Edit Results on Van Gogh houses Dataset Using Different Methods.} Due to page limitations, we have placed some of the experimental results in Appendix~\ref{mr2} and~\ref{mr3} and some comparison results with GPT-4o \cite{gpt4o} in Appendix~\ref{4or}.}
		\label{vg2}
	\end{figure*}
    As shown in Figure~\ref{vg2} (a), a visual comparison of the style transfer results of the \textit{StyleWallfacer} model with other methods is presented. Clearly, the \textit{StyleWallfacer} model achieves the best results in terms of image structure preservation and style transfer. Compared with the results of other methods, the style transfer results of \textit{StyleWallfacer} have truly realized the style transfer, as if the painter himself had redrawn the original image according to his painting style, rather than simply blending the textures and colors of the original and reference images. Moreover, in terms of detail, the results generated by \textit{StyleWallfacer} feature more refined texture details, while other methods generally suffer from noise and damage.
	
	\textbf{One-shot Image-driven Style Transfer \& Color Edit Experimental Qualitative Results.} As depicted in Figure~\ref{vg2} (b), the visualization results of image-driven style transfer and color editing are presented. Analysis of the figure reveals that the proposed method in this paper not only accomplishes style transfer but also retains the model's controllability via text prompts. This enables synchronous guidance of the model's generation process by both "text and style", thereby enhancing controllability. Moreover, the images obtained after style transfer maintain a high degree of content consistency with the original images, achieving a better balance between generation diversity and controllability.
	
	\vspace{-5pt}
	\subsection{Quantitative Comparison}
	\vspace{-5pt}
	\begin{table*}[htbp]
		\centering
		\scalebox{0.52}{
			\begin{tabular}{c|c|c|c|c|c|c|c|c|c|c|c|c}
				\Xhline{1pt}
				\multicolumn{1}{c}{\multirow{2}{*}{Method}} & \multicolumn{4}{|c|}{Landscapes (one-shot)} & \multicolumn{4}{c|}{Van Gogh Houses (one-shot)} & \multicolumn{4}{c}{Watercolor Dogs (one-shot)}  \\ \cline{2-13}
				\multicolumn{1}{c}{}                        & \multicolumn{1}{|>{\columncolor{green!25}}c|}{CLIP-FID ↓} & \multicolumn{1}{>{\columncolor{green!25}}c|}{DINO ↑}  & \multicolumn{1}{>{\columncolor{orange!25}}c|}{CLIP-I ↑}     & \multicolumn{1}{>{\columncolor{purple!25}}c|}{I-LPIPS ↑}      & \multicolumn{1}{|>{\columncolor{green!25}}c|}{CLIP-FID ↓}  & \multicolumn{1}{>{\columncolor{green!25}}c|}{DINO ↑} & \multicolumn{1}{>{\columncolor{orange!25}}c|}{CLIP-I ↑}    & \multicolumn{1}{>{\columncolor{purple!25}}c|}{I-LPIPS ↑}      & \multicolumn{1}{|>{\columncolor{green!25}}c|}{CLIP-FID ↓}  & \multicolumn{1}{>{\columncolor{green!25}}c|}{DINO ↑} & \multicolumn{1}{>{\columncolor{orange!25}}c|}{CLIP-I ↑}    & \multicolumn{1}{>{\columncolor{purple!25}}c|}{I-LPIPS ↑}    \\\Xhline{1pt}
				DreamBooth* \cite{DBLP:conf/cvpr/RuizLJPRA23}                                  &     \cellcolor{blue!25}\textcolor{blue}{29.25}       &   \cellcolor{blue!25}\textcolor{blue}{0.8565}  &  \textcolor{black}{0.8611} & \cellcolor{blue!25}\textcolor{blue}{0.7878}  &\textcolor{black}{28.95} &\textcolor{black}{0.8480} &\textcolor{black}{0.8224} &   \cellcolor{blue!25}\textcolor{blue}{0.7553}         &    \textcolor{black}{35.31} &\textcolor{black}{0.8224} &\textcolor{black}{0.7648} & \textcolor{black}{0.6570}     \\
				DreamBooth+LoRA* \cite{DBLP:conf/iclr/HuSWALWWC22}                            &       \textcolor{black}{29.54}      &  \textcolor{black}{0.8489}& \textcolor{black}{0.8628}&\textcolor{black}{0.7200} &\textcolor{black}{31.08} & \textcolor{black}{0.8316}&\textcolor{black}{0.8000} & \textcolor{black}{0.6611}            &    \textcolor{black}{37.78} &\cellcolor{blue!25}\textcolor{blue}{0.8510} &\cellcolor{blue!25}\textcolor{blue}{0.8124} & \cellcolor{blue!25}\textcolor{blue}{0.7145}         \\
				SVDiff* \cite{DBLP:conf/iccv/HanLZMMY23}                               &            \textcolor{black}{29.53} &\textcolor{black}{0.8406} &\cellcolor{blue!25}\textcolor{blue}{0.8648} & \textcolor{black}{0.7301}&\cellcolor{blue!25}\textcolor{blue}{27.76} &\cellcolor{blue!25}\textcolor{blue}{0.8641} &\cellcolor{blue!25}\textcolor{blue}{0.8642} &   \textcolor{black}{0.7435}       &  \textcolor{black}{45.09} &\textcolor{black}{0.7670} &\textcolor{black}{0.7854} & \textcolor{black}{0.6815}           \\
				Text Inversion* \cite{DBLP:conf/iclr/GalAAPBCC23}                             &   \textcolor{black}{30.58}         &         \textcolor{black}{0.8425}      &             \textcolor{black}{0.8513}  &\textcolor{black}{0.6947} & \textcolor{black}{29.35}&\textcolor{black}{0.8488} &\textcolor{black}{0.8245} &\cellcolor{red!25}\textcolor{red}{0.7616} &         \cellcolor{blue!25}\textcolor{blue}{27.77} &\textcolor{black}{0.8393} &\textcolor{black}{0.7964} & \textcolor{black}{0.6941}         \\
				Ours                                        &       \cellcolor{red!25}\textcolor{red}{28.34} &\cellcolor{red!25}\textcolor{red}{0.8649} &\cellcolor{red!25}\textcolor{red}{0.8712} &\cellcolor{red!25}\textcolor{red}{0.8388} &\cellcolor{red!25}\textcolor{red}{26.44}   & \cellcolor{red!25}\textcolor{red}{0.8649}   &  \cellcolor{red!25}\textcolor{red}{0.8732}   &      \textcolor{black}{0.7084}          &            \cellcolor{red!25}\textcolor{red}{26.64}      &    \cellcolor{red!25}\textcolor{red}{0.8608}           &   \cellcolor{red!25}\textcolor{red}{0.8540}       &        \cellcolor{red!25}\textcolor{red}{0.7205}             \\\Xhline{1pt}
			\end{tabular}
		}
		\caption{\textbf{Quantitative Comparisons to SOTAs on Text-driven Style Transfer Task.} The results that achieve the \textcolor{red}{highest} and \textcolor{blue}{second-highest} performance metrics are respectively delineated in red and blue.} 
		\label{qr}
	\end{table*}

	\begin{table*}[htbp]
		\centering
		\scalebox{0.47}{
			\begin{tabular}{c|c|c|c|c|c|c|c|c|c|c|c|c|c|c|c}
				\Xhline{1pt}
				\multicolumn{1}{c}{\multirow{2}{*}{Method}} & \multicolumn{5}{|c|}{House$\rightarrow$Van Gogh style (text-driven stylization)} & \multicolumn{5}{c|}{House$\rightarrow$Monet style (text-driven stylization)} & \multicolumn{5}{c}{House$\rightarrow$Cezanne style (text-driven stylization)}  \\ \cline{2-13}
				\multicolumn{1}{c}{}                        & \multicolumn{1}{|>{\columncolor{green!25}}c|}{CLIP-FID ↓} & \multicolumn{1}{>{\columncolor{green!25}}c|}{DINO ↑}  &\multicolumn{1}{>{\columncolor{green!25}}c|}{CLIP-I ↑}& \multicolumn{1}{>{\columncolor{purple!25}}c|}{CLIP-T ↑}     & \multicolumn{1}{>{\columncolor{orange!25}}c|}{LPIPS ↓}      & \multicolumn{1}{|>{\columncolor{green!25}}c|}{CLIP-FID ↓}  & \multicolumn{1}{>{\columncolor{green!25}}c|}{DINO ↑} &\multicolumn{1}{>{\columncolor{green!25}}c|}{CLIP-I ↑}& \multicolumn{1}{>{\columncolor{purple!25}}c|}{CLIP-T ↑}    & \multicolumn{1}{>{\columncolor{orange!25}}c|}{LPIPS ↓}      & \multicolumn{1}{|>{\columncolor{green!25}}c|}{CLIP-FID ↓}  & \multicolumn{1}{>{\columncolor{green!25}}c|}{DINO ↑} &\multicolumn{1}{>{\columncolor{green!25}}c|}{CLIP-I ↑}& \multicolumn{1}{>{\columncolor{purple!25}}c|}{CLIP-T ↑}    & \multicolumn{1}{>{\columncolor{orange!25}}c|}{LPIPS ↓}    \\\Xhline{1pt}
				Artist \cite{jiang2024artist}                                  &      \textcolor{black}{70.75}  &  \textcolor{black}{0.7925} &\textcolor{black}{0.6260}&  \textcolor{black}{0.2989} &\textcolor{black}{0.8060} & \cellcolor{blue!25}\textcolor{blue}{68.74}        &\cellcolor{blue!25}\textcolor{blue}{0.6699} &\textcolor{black}{0.4910}&\cellcolor{blue!25}\textcolor{blue}{0.2755}&\textcolor{black}{0.7815}&\textcolor{black}{79.27}        &\cellcolor{blue!25}\textcolor{blue}{0.6587} &\textcolor{black}{0.5302}&\cellcolor{blue!25}\textcolor{blue}{0.2830}&\textcolor{black}{0.7494} \\
				InstructPix2Pix \cite{DBLP:conf/cvpr/BrooksHE23}                            &       \textcolor{black}{72.36}     &\textcolor{black}{0.7464} &  \textcolor{black}{0.5680}  &  \textcolor{black}{0.2378} & \cellcolor{red!25}\textcolor{red}{0.3677}& \textcolor{black}{85.05}        &\textcolor{black}{0.6499} &\textcolor{black}{0.4696}&\textcolor{black}{0.2446}&\cellcolor{blue!25}\textcolor{blue}{0.4135}& \textcolor{black}{77.23}        &\textcolor{black}{0.6424} &\textcolor{black}{0.5336}&\textcolor{black}{0.2693}&\cellcolor{blue!25}\textcolor{blue}{0.4151}         \\
				Plug-and-play \cite{Tumanyan_2023_CVPR}                              &     \cellcolor{blue!25}\textcolor{blue}{57.96}        &\cellcolor{blue!25}\textcolor{blue}{0.7977} &\cellcolor{blue!25}\textcolor{blue}{0.6776}&\cellcolor{blue!25}\textcolor{blue}{0.3086}&\cellcolor{blue!25}\textcolor{blue}{0.4295}&\textcolor{black}{79.87}        &\textcolor{black}{0.6638} &\cellcolor{blue!25}\textcolor{blue}{0.5024}&\textcolor{black}{0.2545}&\cellcolor{red!25}\textcolor{red}{0.2800}     & \cellcolor{blue!25}\textcolor{blue}{73.86}        &\textcolor{black}{0.6576} &\cellcolor{blue!25}\textcolor{blue}{0.5506}&\textcolor{black}{0.2777}&\cellcolor{red!25}\textcolor{red}{0.3295}               \\
				Ours                                        &       \cellcolor{red!25}\textcolor{red}{45.69} &\cellcolor{red!25}\textcolor{red}{0.8075}&\cellcolor{red!25}\textcolor{red}{0.6870}&\cellcolor{red!25}\textcolor{red}{0.3117} & \textcolor{black}{0.7444}& \cellcolor{red!25}\textcolor{red}{57.69}        &\cellcolor{red!25}\textcolor{red}{0.7049} &\cellcolor{red!25}\textcolor{red}{0.5788}&\cellcolor{red!25}\textcolor{red}{0.2859}&\textcolor{black}{0.7268}     & \cellcolor{red!25}\textcolor{red}{63.83}        &\cellcolor{red!25}\textcolor{red}{0.6761} &\cellcolor{red!25}\textcolor{red}{0.5816}&\cellcolor{red!25}\textcolor{red}{0.3145}&\textcolor{black}{0.7042}  \\\Xhline{1pt}
			\end{tabular}
		}
		\caption{\textbf{Quantitative Comparisons to SOTAs on Text-driven Stylization Task.}} 
		\label{qr2}
	\end{table*}
	
	\begin{table*}[htbp]
		\centering
		\scalebox{0.56}{
			\begin{tabular}{c|c|c|c|c|c|c|c|c|c|c|c|c}
				\Xhline{1pt}
				\multicolumn{1}{c|}{\multirow{2}{*}{Method}} & \multicolumn{4}{c|}{Mountain$\rightarrow$Landscapes (one-shot)}     & \multicolumn{4}{c|}{Houses$\rightarrow$Van Gogh Houses (one-shot)}  &\multicolumn{4}{c}{Dogs$\rightarrow$Watercolor Dogs (one-shot)} \\
				\cline{2-13}
				\multicolumn{1}{c|}{}                         & \multicolumn{1}{>{\columncolor{green!25}}c|}{CLIP-FID ↓}&\multicolumn{1}{>{\columncolor{green!25}}c|}{DINO ↑}& \multicolumn{1}{>{\columncolor{green!25}}c|}{CLIP-I ↑} & \multicolumn{1}{>{\columncolor{orange!25}}c|}{LPIPS ↓}  & \multicolumn{1}{>{\columncolor{green!25}}c|}{CLIP-FID ↓}  &\multicolumn{1}{>{\columncolor{green!25}}c|}{DINO ↑}  &\multicolumn{1}{>{\columncolor{green!25}}c|}{CLIP-I ↑}    & \multicolumn{1}{>{\columncolor{orange!25}}c|}{LPIPS ↓}    &\multicolumn{1}{>{\columncolor{green!25}}c|}{CLIP-FID ↓}   &\multicolumn{1}{>{\columncolor{green!25}}c|}{DINO ↑}   &\multicolumn{1}{>{\columncolor{green!25}}c|}{CLIP-I ↑}  & \multicolumn{1}{>{\columncolor{orange!25}}c}{LPIPS ↓}        \\\Xhline{1pt}
				AdaAttn \cite{DBLP:conf/iccv/LiuLHLWLSLD21}                                      &  \textcolor{black}{60.92}  &\textcolor{black}{0.7444} &\textcolor{black}{0.7200} &       \textcolor{black}{0.7613}                           &\textcolor{black}{70.43}   &  \textcolor{black}{0.7839}& \textcolor{black}{0.5825} &   \textcolor{black}{0.7046}         &\textcolor{black}{40.05}     &   \textcolor{black}{0.7455}           &   \textcolor{black}{0.7556}    &     \textcolor{black}{0.6995}               \\
				AdaAIN \cite{DBLP:conf/iccv/HuangB17}                                       &   \textcolor{black}{64.03}   &\textcolor{black}{0.7590} &  \textcolor{black}{0.6942} &   \textcolor{black}{0.7005}                        &\textcolor{black}{73.09}   &  \textcolor{black}{0.7892} & \textcolor{black}{0.5516}  & \textcolor{black}{0.7504}             &\textcolor{black}{37.82}                        &\cellcolor{blue!25}\textcolor{blue}{0.7708}  & \textcolor{black}{0.7530} &     \textcolor{black}{0.7170}            \\
				AesPA-Net \cite{DBLP:conf/iccv/HongJLAKLKUB23}                                      & \textcolor{black}{61.71}  &\textcolor{black}{0.7554}& \textcolor{black}{0.6996}&    \textcolor{black}{0.6592}                        &\textcolor{black}{65.65} &  \textcolor{black}{0.7887}& \textcolor{black}{0.5979}&        \textcolor{black}{0.7380}    & \textcolor{black}{39.92}                        & \textcolor{black}{0.7438}& \cellcolor{blue!25}\textcolor{blue}{0.7645} &     \cellcolor{blue!25}\textcolor{blue}{0.6677}                    \\
				StyleID \cite{DBLP:conf/cvpr/ChungHH24}                                     & \cellcolor{blue!25}\textcolor{blue}{47.45} & \textcolor{black}{0.7518} & \cellcolor{blue!25}\textcolor{blue}{0.7564} & \textcolor{black}{0.6062}                    &      \cellcolor{blue!25}\textcolor{blue}{55.79}  & \cellcolor{blue!25}\textcolor{blue}{0.7996} & \cellcolor{blue!25}\textcolor{blue}{0.6501} & \cellcolor{blue!25}\textcolor{blue}{0.7183} &  \cellcolor{blue!25}\textcolor{blue}{36.83}                    &   \textcolor{black}{0.7615}    & \textcolor{black}{0.7613}    &     \textcolor{black}{0.6859}             \\
				InstantStyle-Plus \cite{DBLP:journals/corr/abs-2407-00788}                                      &  \textcolor{black}{59.04}   &\cellcolor{blue!25}\textcolor{blue}{0.7595} & \textcolor{black}{0.7381}&      \cellcolor{red!25}\textcolor{red}{0.3909} &\textcolor{black}{64.32}   & \textcolor{black}{0.7582}& \textcolor{black}{0.6077} &    \cellcolor{red!25}\textcolor{red}{0.2903}      &\textcolor{black}{41.04}      &      \textcolor{black}{0.7437}      &     \textcolor{black}{0.7633}    &    \cellcolor{red!25}\textcolor{red}{0.3132}                    \\
				Ours                                         &\cellcolor{red!25}\textcolor{red}{45.14}   &\cellcolor{red!25}\textcolor{red}{0.8124}  & \cellcolor{red!25}\textcolor{red}{0.8210} & \cellcolor{blue!25}\textcolor{blue}{0.5917} &\cellcolor{red!25}\textcolor{red}{37.19}  & \cellcolor{red!25}\textcolor{red}{0.8346} & \cellcolor{red!25}\textcolor{red}{0.7309}  &       \textcolor{black}{0.7437}                   &\cellcolor{red!25}\textcolor{red}{35.40}                          &\cellcolor{red!25}\textcolor{red}{0.8041} & \cellcolor{red!25}\textcolor{red}{0.7852} &   \textcolor{black}{0.6848}             \\ \Xhline{1pt}
			\end{tabular}
		}
		\caption{\textbf{Quantitative Comparisons to SOTAs on Image-driven Style Transfer Task.}} 
		\label{abqr}
	\end{table*}
	As shown in Table~\ref{qr}, Table~\ref{qr2} and Table~\ref{abqr}, the method proposed in this paper achieved the best results compared with all the baseline methods, further demonstrating the effectiveness of the proposed method from a quantitative perspective.

	\subsection{Ablation Study}
	\label{abl}
	To prove that the proposed techniques can indeed effectively improve the performance of \textit{StyleWallfacer} in various generation scenarios, we conduct extensive ablation studies focusing on these techniques and leave them in Appendix~\ref{abla} due to page limit. And we have also understood the source of \textit{StyleWallfacer}'s superiority from a mathematical perspective, for details see Appendix~\ref{math}.
	\section{Conclusion}
	In this work, we focus on building a unified framework for style transfer by analyzing semantic drift, overfitting, and the true meaning of style transfer that previous works have failed to settle, and accordingly proposing a new method named \textit{StyleWallfacer}. \textit{StyleWallfacer} includes a one-stage fine-tuning process and a training-free inference framework that aims to solve these issues, namely the semantic-based style learning strategy, the training-free triple diffusion process, and the data augmentation method for small scale datasets based on human feedback. With these designs tailored to style transfer, our \textit{StyleWallfacer} achieves convincing performance on text/image-driven style transfer scenarios, text-driven stylization, and image-driven style transfer with color edit, while solving problems before. In Appendix~\ref{LF} and~\ref{BI}, we will discuss possible limitations and potential future works of \textit{StyleWallfacer}.
	
	\clearpage
	\bibliographystyle{plain}
	\bibliography{ref}
	
	\clearpage
	\appendix
	
	\section*{Appendix}
	\subsection*{Overview}
	This supplementary material provides the relsted works, additional experiments and results to further support our main findings and proposed \textit{StyleWallfacer}.
	These were not included in the main paper due to the space limitations.
	The supplementary material is organized as follows:	

	{
		\hypersetup{linkcolor=blue}
		\tableofcontents
	}
	\clearpage
	
	\section{Related Works}
	\label{gen_inst}
	\subsection{One-shot Text-driven Style Transfer}

	\subsection{One-shot Image-driven Style Transfer}
   \subsection{Text-driven Image Stylization}
   \section{Implementation Detail}
	\label{EXP}
	\subsection{Model}
\subsection{Training}
		\subsection{Inference}
	\label{inf}

	\subsection{Details about the LLM}
	
	\section{The Mathematical Explanation of the Efficacy of \textit{StyleWallfacer}}

	\subsection{Study on the Impact of Noise Time Thresholds on Model Generation Outcomes}

	\subsubsection{Study on the \( t_s^{l} \)}

	\subsubsection{Study on the \( t_s^{s} \)}

	\subsection{Generalizability Study on Image Edit}
	\section{Problems of Existing Methods and Their Visualizations}
	
	\subsection{Limited Color Domain}
	\label{LCD}
\subsection{Risk of Overfitting}
	\label{RO}
\section{Additional Analysis}
	\subsection{Comparison with GPT-4o \cite{gpt4o} Image Generation}
	\label{4or}
\section{Additional Results}
	\label{mr}
	\subsection{One-shot Text-driven Style Transfer}
	\label{mr1}
\subsection{Text-driven Styliztion}
	\label{mr4}
\subsection{One-shot Image-driven Style Transfer}
	\label{mr2}

	\subsection{One-shot Text-driven Style Transfer \& Color Edit}
	
	\label{mr3}

	\subsection{Cross-Content Image Testing Results}
	\section{Limitation and Future Work}
	\label{LF}

	\section{Broader Impact}
	\label{BI}

\end{document}